\documentclass[a4paper]{article}
\usepackage{CJKutf8}
\usepackage{algorithm}
\usepackage{algorithmic}
\usepackage{xcolor}
\usepackage[utf8]{inputenc}
\usepackage[most]{tcolorbox}
\usepackage{xcolor}
\usepackage{fontawesome5}
\usepackage{enumitem}
\usepackage{amssymb}

\usepackage{geometry}
\usepackage{makecell}
\usepackage{adjustbox}

\definecolor{lightgray}{gray}{0.95}
\definecolor{spk0}{HTML}{1f77b4} 
\definecolor{spk1}{HTML}{2ca02c} 

\usepackage{INTERSPEECH2022}

\title{A Unified Speech LLM for Diarization and Speech Recognition in \\Multilingual Conversations}
\name{Phurich Saengthong$^1$, Boonnithi Jiaramaneepinit$^2$, Sheng Li$^1$, \\ Manabu Okumura$^2$, Takahiro Shinozaki$^1$}
\address{
  Institute of Science Tokyo}
\email{www.ts.ip.titech.ac.jp$^1$, www.lr.first.iir.isct.ac.jp$^2$}

\begin{document}

\maketitle
\begin{abstract}
Speech Large Language Models (Speech LLMs) have emerged as a crucial paradigm in recent years, extending the capabilities of traditional LLMs to speech tasks such as automatic speech recognition (ASR) and spoken dialogue modeling. However, their effectiveness in real-world multilingual conversations remains limited by the scarcity of data that captures natural conversational phenomena. To address this, the MLC-SLM Challenge provides a multilingual conversational dataset and evaluates models on two tasks: ASR with oracle segmentation (Task I) and joint diarization and recognition without oracle information (Task II). In this paper, we focus on Task II and propose a unified speech LLM that jointly performs diarization and ASR in an end-to-end manner. By reformulating the training data format and modifying the inference procedure, our model addresses the ambiguity inherent in pre-segmented audio and achieves a 54.87\% relative improvement in tcpWER/tcpCER over the baseline, ranking 8th overall, despite using a smaller LLM backbone. We also report results from Task I using a fine-tuned speech LLM.
\end{abstract}

\vspace{-3mm}
\section{Introduction}
Large Language Models (LLMs) have recently been extended to speech tasks such as automatic speech recognition (ASR) and spoken dialogue modeling. However, progress remains constrained by the scarcity of real-world conversational data—especially in multilingual contexts—where complex patterns like speaker overlaps, interruptions, and diverse speaking styles are common. To address this gap, the MLC-SLM Challenge and Workshop aim to advance multilingual conversational speech language modeling by providing a new real-world multilingual conversational speech dataset \cite{nexdata2025mlc_slm}.

The challenge comprises two tracks that assess different capabilities of speech language models. Task I targets multilingual ASR, where systems transcribe speech using oracle speaker labels and segmentation, and are evaluated by Word Error Rate (WER) or Character Error Rate (CER). Task II focuses on joint speaker diarization and recognition without oracle information, requiring systems to both identify speaker turns and transcribe speech. Evaluation metrics for Task II include Diarization Error Rate (DER) and tcpWER/tcpCER, which jointly reflect diarization and transcription accuracy.

In this paper, we focus on Task II. We propose a unified speech LLM that performs diarization and ASR in an end-to-end fashion, overcoming a key limitation of the baseline system—its inability to handle ambiguous audio inputs caused by pre-segmentation. By reformulating the training data format and modifying the inference procedure, our approach achieved a 54.87\% relative improvement in tcpWER/tcpCER over the official baseline, despite using a smaller backbone model (3B vs. 8B). Additionally, we report our results from Task I, obtained by fine-tuning an existing speech LLM.

Our key contributions are as follows:
\begin{itemize}
\item We present a unified speech LLM that jointly performs diarization and ASR, effectively addressing ambiguity in audio that the baseline system fails to resolve.
\item Our model achieved a 54.87\% relative improvement over the Task II baseline in tcpWER/tcpCER, placing 8th in the official ranking.
\item We additionally report our Task I results using a fine-tuned speech LLM, which achieved 20th place.
\end{itemize}

\vspace{-2mm}


\begin{figure*}[!t]
    \centering
    \includegraphics[width=0.85\textwidth]{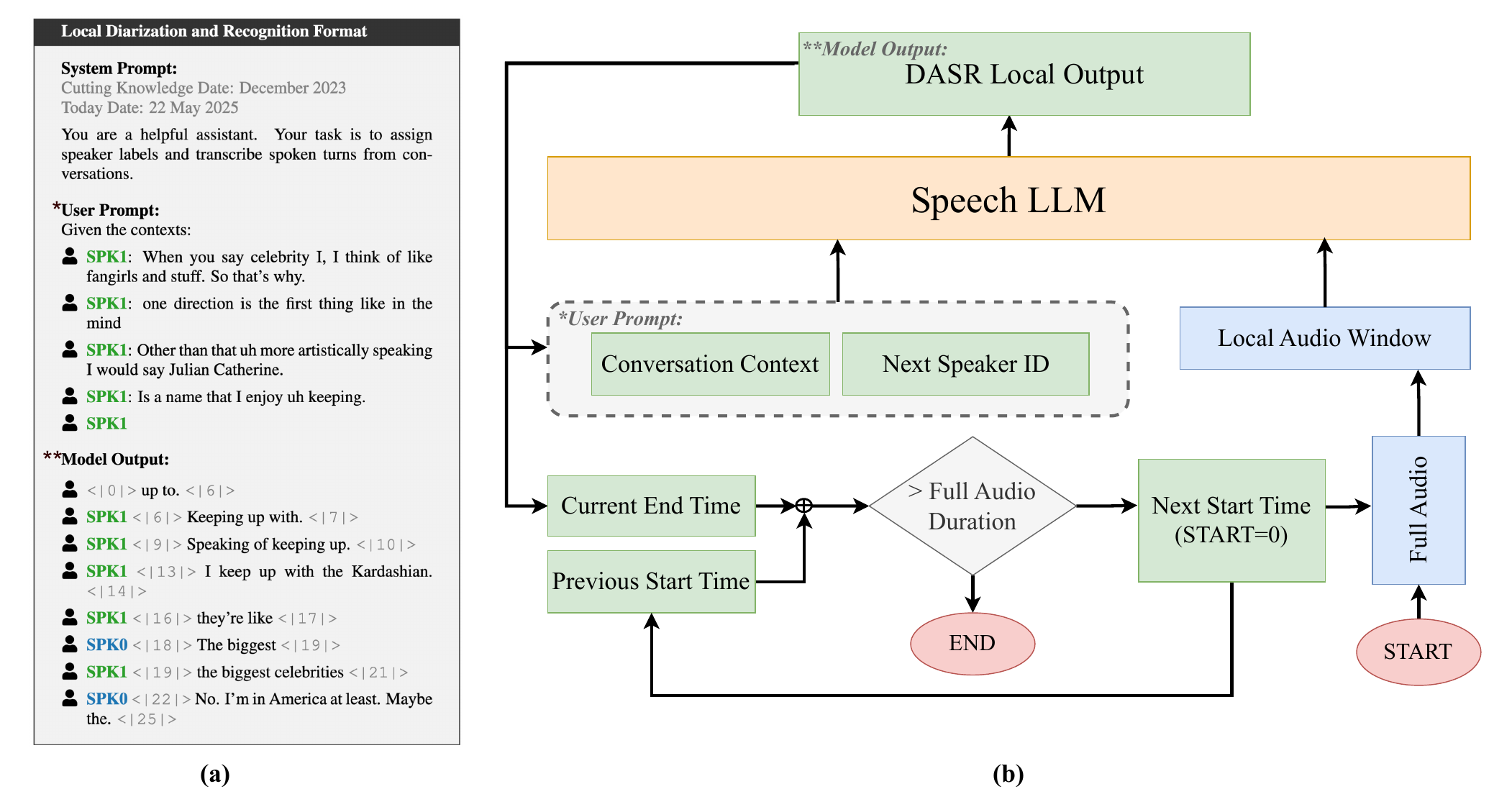}
    \caption{Overview of our approach for local diarization and speech recognition using Speech LLM. (a) Example input-output format, where the model receives a system prompt, an optional user prompt containing recent dialogue history, and produces diarized transcriptions with speaker labels and timestamps. (b) Inference is performed iteratively on local audio windows, using the updated speaker context and predicted next speaker to maintain continuity until the full audio is processed.}
    \label{fig:task2_method}
    \vspace{-4mm}
\end{figure*}

\section{End-to-end Speech LLM for Joint Diarization and ASR (Task II)}



In Task II, the baseline system \cite{mlcslm2025baseline} performs diarization and recognition in separate stages: it applies voice activity detection (VAD) and speaker clustering to segment the audio, then uses a LLM-based ASR model to transcribe each segment independently \cite{nexdata2025mlc_slm}. This separation often lacks sufficient conversational context, making it difficult to resolve ambiguous speaker transitions or maintain consistent speaker assignments, especially in multilingual, overlapping, or turn-taking scenarios.

To address these limitations, we propose an end-to-end approach for joint speaker diarization and automatic speech recognition (DASR) using a speech large language model (Speech LLM). Our method combines a locally sliding (non-overlapping) inference window with prompt-based context from prior speaker turns to jointly predict who is speaking and what is being said (Figure~\ref{fig:task2_method}). The model is trained with a Local Diarization and Recognition format, enabling it to learn structured speaker-text alignment. At inference time, it processes the conversation iteratively over local windows using updated speaker context and predicted speaker prompts for continuity. While related to prior end-to-end DASR models~\cite{onemodel, park2024sortformerseamlessintegrationspeaker}, our work uniquely employs an LLM as the core backbone for diarization-aware transcription.

\vspace{-1mm}
\subsection{Speech LLM}
The architecture of our Speech LLM closely mirrors that of the baseline system \cite{mlcslm2025baseline}, with the key difference being our use of \texttt{Llama-3.2-3B-instruct} \cite{grattafiori2024llama3herdmodels} \cite{meta_llama-3.2-3b-instruct} as the LLM backbone, replacing the \texttt{Llama-3.1-8B} \cite{meta_llama-3.1-8b} model. The architecture of our Speech LLM consists of four main components: a speech encoder, a sub-sampling projector, a learnable prompt, and an LLM enhanced with LoRA \cite{hu2021lora}. We adopt the Whisper encoder \cite{radford2022robustspeechrecognitionlargescale}, which is built upon Transformer layers \cite{transformer}, to extract rich speech representations. The sub-sampling projector includes four layers: two 1D convolutional layers with a GELU \cite{gelu} activation in between to reduce the temporal resolution, followed by two linear layers with a ReLU \cite{agarap2019deeplearningusingrectified} activation to project features into the target LLM dimension, followed by a LayerNorm \cite{ba2016layernormalization}. For the LLM, we use \texttt{Llama-3.2-3B-instruct} with LoRA adaptation. 

\vspace{-1mm}
\subsection{Local Diarization and Recognition Format}

In real-world conversations, speaker turns are often ambiguous, making it difficult for systems that treat diarization and transcription as separate tasks. The baseline system \cite{mlcslm2025baseline} follows this separation, which limits its ability to resolve unclear speaker transitions or leverage contextual cues across modalities \cite{nexdata2025mlc_slm}. We hypothesize that while larger Speech LLMs have the potential to improve reasoning and language understanding, they remain suboptimal if they lack the ability to interpret ambiguous speaker turns and conversational structure. 

As illustrated in Figure~\ref{fig:task2_method}a, we introduce a unified data format that enables a (smaller) Speech LLM to jointly model both who is speaking and what is being said. This format interleaves speaker tokens and timestamp tokens with transcribed text, allowing the model to generate speaker-labeled utterances with temporal alignment. It also supports conditioning on recent transcript history, enabling the model to leverage long-range textual context to disambiguate speech—particularly when speaker identity is unclear from audio alone. Optional user prompts can further guide the model’s understanding of speaker roles and dialogue flow.

We use a fixed set of special timestamp tokens to mark the start and end of each speaker segment, similar to the design of Whisper~\cite{radford2022robustspeechrecognitionlargescale} \cite{onemodel}. The model is designed for two-speaker conversations and uses two speaker tokens: \texttt{<|SPK0|>} and \texttt{<|SPK1|>}. By default, the model assigns \texttt{<|SPK0|>} to the first detected speaker in the audio. If the user provides an initial speaker prompt, such as \texttt{<|SPK1|>}, the model respects that choice and automatically assigns the other token, \texttt{<|SPK0|>}, to the remaining speaker as the conversation unfolds. This mechanism supports flexible prompting while maintaining consistent speaker attribution. The model is trained and evaluated using both prompt-based and prompt-free formats to handle both contextualized and contextless conversation starts.

\vspace{-1mm}
\subsection{Training Details}
We train the Speech LLM on the multilingual dataset provided by the MLC-SLM challenge, which consists of two-speaker conversations across eleven languages, including five English accents. The speaker (\texttt{<|SPK0|>}, \texttt{<|SPK1|>}) and timestamp tokens are registered as trainable special tokens and integrated into the model’s tokenizer and embedding table. To support this, we further fine-tune the embedding layer so the model can effectively learn the representation and usage of these tokens in context. During training, the Whisper encoder and the rest of the LLM backbone are kept frozen; only the sub-sampling projector, prompt embeddings, LoRA parameters, and the embedding layer (including special token embeddings) are updated to reduce memory usage and computational overhead.

We adopt the baseline’s default hyperparameters, applying SpecAugment \cite{park2019specaugment} with two frequency masks of width 10 and two time masks of width 50, SpecSubstitute \cite{wu2021u2unifiedtwopassbidirectional} with three time substitutions of width 30, and speed perturbation. Each batch is dynamically limited to a maximum of 8000 frames. The model is trained using Adam with an initial learning rate of 0.0001 and a warmup-based scheduler with 2500 warmup steps, for a total of 130,000 iterations on a single RTX 6000 GPU.

\vspace{-1mm}
\subsection{Inference Details} 
\subsubsection{Local Diarization and Speech Recognition}
To perform offline diarization and transcription, our system processes the audio iteratively in segments. Rather than dividing the audio into fixed, equally spaced windows, each new segment begins at a time determined by the previous output, ensuring alignment with meaningful conversational boundaries. Specifically, a fixed chunk duration of 24 seconds defines the maximum context window that the Speech LLM operates on during each forward pass. However, the actual start time of each segment is dynamically updated by taking the previous chunk’s start time and adding the end time of the second-to-last predicted speaker turn from that chunk. This approach avoids artificial segmentation and ensures that the model resumes transcription from a coherent point in the conversation. The entire process is performed by a single Speech LLM. The overall inference loop is illustrated in Figure~\ref{fig:task2_method}b.

For the first segment, the model is prompted with a general system instruction, with no user-provided speaker prompt or dialogue history. For all subsequent segments, the prompt includes a sliding context window consisting of the four most recent speaker turns from previous chunks, along with the second-to-last predicted speaker token, which is used to guide the model in assigning speaker labels consistently across the session. This mechanism enables the model to carry contextual information forward and resolve ambiguity in turn-taking.

If the model produces no valid turns in a segment—such as in cases of silence or hallucination, which can be detected when no speaker label or end time is returned—a small time offset of 0.3 seconds is added to the current endpoint to advance the inference window. This iterative process enables the model to produce diarized and temporally coherent transcripts without requiring access to the entire audio file at once or relying on overlapping segments.

\begin{table}[t]
\centering
\caption{An official result of MLC-SLM Task II on the evaluation set. tcpWER/tcpCER is evaluated with a 5\,s collar. Evaluation is conducted using the \texttt{MeetEval} toolkit~\cite{vonneumann2023meeteval}.}
\begin{tabular}{c l c}
\toprule
\textbf{Rank} & \textbf{Participant} & \textbf{tcpWER/tcpCER} \\
\midrule
1 & MegaAIS                   & 16.53 \\
2 & but\_mlc                  & 16.75 \\
3 & tenp1                    & 17.49 \\
4 & seewo                    & 17.67 \\
5 & Duke Kunshan University & 18.08 \\
6 & sixteen-years           & 19.27 \\
7 & llm-t                    & 26.30 \\
\midrule
8 & ST-ShinozakiLab         & 27.25 \\
\midrule
9 & fosafer                  & 31.68 \\
10 & voicecode                & 55.96 \\
11 & 517517                   & 59.40 \\
12 & mlcslm-baseline \cite{mlcslm2025baseline}        & 60.39 \\
\bottomrule
\end{tabular}
\label{tab:eval_task2}
\vspace{-5mm}
\end{table}

\subsubsection{Local and Global Diarization Alignment}

Although the Speech LLM can perform diarization locally within each audio segment, it lacks global context, which often leads to incorrect speaker assignments in later turns and results in cascading errors across a conversation. These inconsistencies degrade speaker coherence and contribute to a higher diarization error rate (DER) over full audio.

To address this, we apply a global diarization module used in the baseline system \cite{mlcslm2025baseline} to obtain RTTM files containing speaker segments with consistent identities across the full audio. We then design a post-processing module that aligns the locally predicted speaker turns from the Speech LLM (in STM format) with the RTTM segments. For each STM segment, the module identifies overlapping RTTM speaker segments and assigns the speaker label with the greatest temporal overlap, grounding the speaker identity in a more stable diarization output.

Adjacent STM segments that share the same assigned speaker and have nearly identical start or end times (within 0.01 seconds) are merged to form a continuous, coherent turn. If no RTTM overlap is found, the original LLM-predicted speaker label and timestamps are retained.


\begin{table}[t]
\centering
\caption{Model comparison for MLC-SLM Task II on the development set. All systems use the same encoder (Whisper-large-v3) and diarization module (3D-Speaker). DER is evaluated with a 0.25\,s collar, and tcpWER/tcpCER with a 5\,s collar. \textcolor{gray}{*For simplicity, we report orcWER in place of WER.} Evaluation is conducted using the \texttt{MeetEval} toolkit~\cite{vonneumann2023meeteval}.}
\resizebox{\columnwidth}{!}{%
\begin{tabular}{l c c c c c}
\toprule
\textbf{Model} & \textbf{LLM} & \textbf{DER} & \textbf{WER} & \textbf{cpWER} & \textbf{tcpWER/tcpCER} \\
\midrule
Baseline & \texttt{Llama-3.1-8B} \cite{meta_llama-3.1-8b} & 16.44 & 21.56 & --    & 76.12 \\
Ours     & \texttt{Llama-3.2-1B-instruct} \cite{meta_llama-3.2-1b-instruct} & 14.03 & \textcolor{gray}{23.88*} & 28.24 & 31.19 \\
Ours     & \texttt{Llama-3.2-3B-instruct} \cite{meta_llama-3.2-3b-instruct} & 13.36 & \textcolor{gray}{20.92*} & 24.97 & 28.23 \\
\bottomrule
\end{tabular}
}
\label{tab:llm-comparison}
\vspace{-5mm}
\end{table}

\vspace{-2mm}
\subsection{Results}

Table~\ref{tab:eval_task2} shows that our system achieved 8\textsuperscript{th} place in Task II: Multilingual Conversational Speech Diarization and Recognition. Our approach achieved significantly lower tcpWER/tcpCER than the baseline system~\cite{mlcslm2025baseline}, which was consistent with the results on the development dataset, as shown in Table~\ref{tab:llm-comparison}. Specifically, our system achieved a tcpWER/tcpCER of 27.25 on the evaluation set and 28.23 on the development set, compared to 60.39 and 76.12, respectively, for the baseline.

In terms of diarization quality, our method reduced the Diarization Error Rate (DER) from 16.44\% in the baseline to 14.03\% using our 1B model, and further to 13.36\% with our 3B model, demonstrating consistent improvements in speaker attribution. These gains are attributed to the integration of diarization outputs from the 3D-Speaker system \cite{wang2023campp} with our Speech LLM, which jointly models speaker identity and transcription in context.

\begin{figure}[!t]
    \centering
    \includegraphics[width=\linewidth]{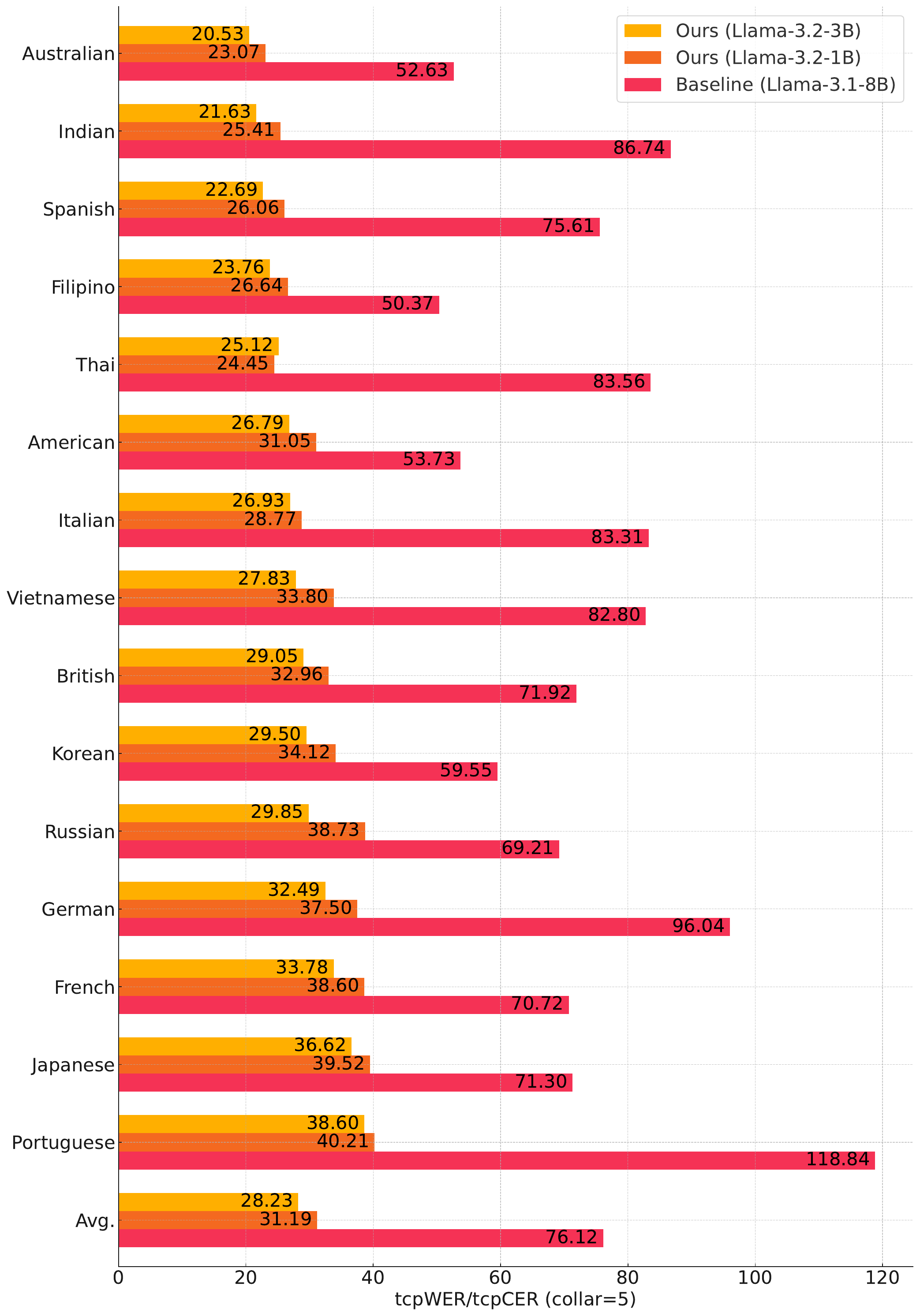}
    \caption{Per-language tcpWER/tcpCER (\%) comparison on the development set for MLC-SLM Task II, between the baseline system \cite{mlcslm2025baseline} (\texttt{Llama-3.1-8B}) and our models using Llama-3.2-1B and Llama-3.2-3B. All models use the same encoder (Whisper-large-v3) and diarization system (3D-Speaker). Lower is better. Evaluation is conducted using the \texttt{MeetEval} toolkit~\cite{vonneumann2023meeteval}.}
    \label{fig:lid_tcpWER_comparison}
    \vspace{-6mm}
\end{figure}

In addition, we evaluated a smaller version of our model, replacing \texttt{Llama-3.2-3B-instruct} with \texttt{Llama-3.2-1B-instruct} while keeping all training settings identical. Table~\ref{tab:llm-comparison} and Figure~\ref{fig:lid_tcpWER_comparison} compare the performance across the two LLM sizes and the baseline. We found that the smaller LLM version resulted in a higher tcpWER by 2.96\%. Interestingly, Thai was the only language where the smaller model slightly outperformed the larger model (24.45\% vs.\ 25.12\%), potentially due to domain variability or the overfitting sensitivity of the larger model.

Overall, our approach achieved a 54.87\% relative improvement in tcpWER/tcpCER compared to the baseline, as shown in Table~\ref{tab:eval_task2}, demonstrating the effectiveness of our locally structured diarization and recognition method—even when using a much smaller LLM (3B vs.\ 8B).

\section{Fine-tuning ASR-based Speech LLM (Task I)}
In this section, we report the additional experiment for Task I. Specifically, our primary goal was to assess the effectiveness of fine-tuning a multilingual multimodal model under a two-phase strategy that accommodates both language-specific and cross-lingual characteristics. Given the diverse linguistic and acoustic properties across target languages, it was essential to design an experimental pipeline that maintained both language-level adaptation and multilingual generalization. This approach was intended to enhance the model's performance, particularly on underrepresented languages and more challenging audio-text alignment scenarios.

\subsection{Training Details}

We fine-tuned the \texttt{Phi-4-multimodal-instruct} model in two stages. In the first phase, Language-Specific Pretraining, we trained the model separately on each language using data segmented into chunks of up to 90 seconds. This enabled the model to capture language-specific phonetic and syntactic features effectively. In the second phase, Unified Multilingual Fine-Tuning, we trained the model on a merged multilingual dataset, with chunking based on the actual timestamp boundaries in the audio-text pairs. This allowed for better preservation of natural utterance structure and facilitated the model’s cross-lingual generalization. The \texttt{Phi-4-multimodal-instruct} model contains a total of 5.6 billion parameters, of which 830,472,192 parameters were fine-tuned in our training pipeline, enabling the model to adapt both to individual linguistic traits and shared multilingual patterns effectively \cite{phi4multimodal}.

\textbf{Hyperparameter Configuration.} Training used the AdamW optimizer \cite{loshchilov2017decoupled} with fixed hyperparameters: $\beta_1 = 0.9$, $\beta_2 = 0.95$, $\epsilon = 10^{-7}$, and a learning rate of $2 \times 10^{-5}$. A linear schedule with 50 warm-up steps, gradient clipping (max norm 1.0), and weight decay of 0.01 were applied. These settings followed standard multilingual transformer fine-tuning practices and ensured stable convergence \cite{kim2025phi4mm_ko_stt}.

\begin{table}[t]
\centering
\caption{Unified results for MLC-SLM Task I: official rankings and language-specific WER/CER at 1 and 2.5 epochs. Evaluation is conducted using the \texttt{MeetEval} toolkit~\cite{vonneumann2023meeteval}.}
\label{tab:unified_results}
\begin{adjustbox}{width=\columnwidth,center}
\begin{tabular}{@{}c l c c @{}}
\toprule
\multicolumn{4}{l}{\textbf{Official Evaluation Results}} \\
\midrule
\textbf{Rank} & \textbf{Participant} & \multicolumn{2}{l}{\textbf{WER/CER (\%)}} \\
\midrule
1  & sixteen-years      & 8.88  & -- \\
2  & tenp1              & 9.60  & -- \\
3  & t-asr              & 9.83  & -- \\
18 & mlcslm-baseline \cite{mlcslm2025baseline}    & 20.17 & -- \\
20 & ST-ShinozakiLab (@1ep)        & 25.56 & -- \\
\midrule
\multicolumn{4}{l}{\textbf{Development Results}} \\
\midrule
\multicolumn{2}{l}{\textbf{Per-Language}} 
& \makecell{\textbf{WER/CER} \\ \textbf{@1ep (\%)}} 
& \makecell{\textbf{WER/CER} \\ \textbf{@2.5ep (\%)}} \\
\midrule
\multicolumn{2}{l}{Filipino (English)}     & 8.94  & 8.66  \\
\multicolumn{2}{l}{Spanish}                & 9.76  & 19.00 \\
\multicolumn{2}{l}{British (English)}      & 11.44 & 13.49 \\
\multicolumn{2}{l}{American (English)}     & 12.31 & 16.46 \\
\multicolumn{2}{l}{Australian (English)}   & 12.31 & 24.31 \\
\multicolumn{2}{l}{Italian}                & 14.05 & 24.67 \\
\multicolumn{2}{l}{Korean}                 & 16.57 & 15.63 \\
\multicolumn{2}{l}{German}                 & 19.35 & 18.77 \\
\multicolumn{2}{l}{French}                 & 20.57 & 22.22 \\
\multicolumn{2}{l}{Japanese}               & 20.88 & 19.46 \\
\multicolumn{2}{l}{Indian (English)}       & 25.06 & 15.46 \\
\multicolumn{2}{l}{Portuguese}             & 28.05 & 30.48 \\
\multicolumn{2}{l}{Thai}                   & 49.47 & 36.21 \\
\multicolumn{2}{l}{Vietnamese}             & 64.86 & 56.30 \\
\multicolumn{2}{l}{Russian}                & 73.60 & 57.21 \\
\midrule
\multicolumn{2}{l}{\textbf{Average}}       & \textbf{27.81} & \textbf{25.21} \\
\bottomrule
\end{tabular}
\end{adjustbox}
\vspace{-6mm}
\end{table}

\subsection{Results}

Table~\ref{tab:unified_results} (upper part) shows the official results from the MLC-SLM Task I evaluation set, using oracle segmentation and speaker labels. Our submission ("ST-ShinozakiLab @1ep") achieved a WER/CER of 25.56\%, ranking 20\textsuperscript{th}, while the baseline system reached 20.17\%. Due to significant computational and time constraints, our model was trained for only one epoch, which limited its performance.

After the challenge \cite{nexdata2025mlc_slm}, we extended training to 2.5 epochs to assess the impact of longer training. The lower part of Table~\ref{tab:unified_results} presents WER/CER results across multiple languages for models trained at 1 and 2.5 epochs. In many cases, performance improved. For example, English-Indian WER dropped from 25.06\% to 15.46\%, and Filipino-English also saw a slight improvement. However, certain languages, such as Australian English and Spanish, showed notable degradation. 

\section{Conclusions}

In this work, we proposed a unified speech LLM for joint diarization and ASR in multilingual conversations, achieving a 54.87\% relative improvement in tcpWER/tcpCER over the Task II baseline and ranking 8th with a smaller 3B model. 
For Task I, our minimally trained system ranked 20th, but post-challenge training to 2.5 epochs showed notable gains in several languages, highlighting the model’s potential and the need for further optimization.

\section{Acknowledgements}

The authors thank Haoyuan Yang, Pintusorn Suttiponpisarn, and Sakulthip Rassameecharerntham for valuable discussions on related work.

\bibliographystyle{IEEEtran}

\bibliography{mybib}

\begin{thebibliography}{10}
\providecommand{\url}[1]{#1}
\csname url@samestyle\endcsname
\providecommand{\newblock}{\relax}
\providecommand{\bibinfo}[2]{#2}
\providecommand{\BIBentrySTDinterwordspacing}{\spaceskip=0pt\relax}
\providecommand{\BIBentryALTinterwordstretchfactor}{4}
\providecommand{\BIBentryALTinterwordspacing}{\spaceskip=\fontdimen2\font plus
\BIBentryALTinterwordstretchfactor\fontdimen3\font minus \fontdimen4\font\relax}
\providecommand{\BIBforeignlanguage}[2]{{%
\expandafter\ifx\csname l@#1\endcsname\relax
\typeout{** WARNING: IEEEtran.bst: No hyphenation pattern has been}%
\typeout{** loaded for the language `#1'. Using the pattern for}%
\typeout{** the default language instead.}%
\else
\language=\csname l@#1\endcsname
\fi
#2}}
\providecommand{\BIBdecl}{\relax}
\BIBdecl

\bibitem{nexdata2025mlc_slm}
\BIBentryALTinterwordspacing
``{MLC-SLM Challenge}: Multilingual conversational speech language model at interspeech 2025,'' Online competition and workshop, Nexdata, Aug. 2025, hosted satellite workshop at Interspeech 2025. [Online]. Available: \url{https://www.nexdata.ai/competition/mlc-slm}
\BIBentrySTDinterwordspacing

\bibitem{mlcslm2025baseline}
M.~Shen, ``Mlc-slm baseline,'' \url{https://github.com/mubingshen/MLC-SLM-Baseline/tree/main}, 2025.

\bibitem{onemodel}
S.~Cornell, J.-W. Jung, S.~Watanabe, and S.~Squartini, ``One model to rule them all ? towards end-to-end joint speaker diarization and speech recognition,'' in \emph{ICASSP 2024 - 2024 IEEE International Conference on Acoustics, Speech and Signal Processing (ICASSP)}, 2024, pp. 11\,856--11\,860.

\bibitem{park2024sortformerseamlessintegrationspeaker}
\BIBentryALTinterwordspacing
T.~Park, I.~Medennikov, K.~Dhawan, W.~Wang, H.~Huang, N.~R. Koluguri, K.~C. Puvvada, J.~Balam, and B.~Ginsburg, ``Sortformer: Seamless integration of speaker diarization and asr by bridging timestamps and tokens,'' 2024. [Online]. Available: \url{https://arxiv.org/abs/2409.06656}
\BIBentrySTDinterwordspacing

\bibitem{grattafiori2024llama3herdmodels}
\BIBentryALTinterwordspacing
A.~Grattafiori \emph{et~al.}, ``The llama 3 herd of models,'' 2024. [Online]. Available: \url{https://arxiv.org/abs/2407.21783}
\BIBentrySTDinterwordspacing

\bibitem{meta_llama-3.2-3b-instruct}
Meta, ``Llama-3.2-3b-instruct,'' \url{https://huggingface.co/meta-llama/Llama-3.2-3B-Instruct}, 2024.

\bibitem{meta_llama-3.1-8b}
------, ``Llama-3.1-8b,'' \url{https://huggingface.co/meta-llama/Llama-3.1-8B}, 2024.

\bibitem{hu2021lora}
E.~J. Hu, Y.~Shen, P.~Wallis, Z.~Allen‑Zhu, Y.~Li, S.~Wang, L.~Wang, and W.~Chen, ``Lora: Low‑rank adaptation of large language models,'' in \emph{Proceedings of the International Conference on Learning Representations (ICLR)}, 2022, arXiv:2106.09685.

\bibitem{radford2022robustspeechrecognitionlargescale}
\BIBentryALTinterwordspacing
A.~Radford, J.~W. Kim, T.~Xu, G.~Brockman, C.~McLeavey, and I.~Sutskever, ``Robust speech recognition via large-scale weak supervision,'' 2022. [Online]. Available: \url{https://arxiv.org/abs/2212.04356}
\BIBentrySTDinterwordspacing

\bibitem{transformer}
A.~Vaswani, N.~Shazeer, N.~Parmar, J.~Uszkoreit, L.~Jones, A.~N. Gomez, L.~Kaiser, and I.~Polosukhin, ``Attention is all you need,'' in \emph{Proceedings of the 31st International Conference on Neural Information Processing Systems}, ser. NIPS'17.\hskip 1em plus 0.5em minus 0.4em\relax Red Hook, NY, USA: Curran Associates Inc., 2017, p. 6000–6010.

\bibitem{gelu}
\BIBentryALTinterwordspacing
D.~Hendrycks and K.~Gimpel, ``Bridging nonlinearities and stochastic regularizers with gaussian error linear units,'' 2016. [Online]. Available: \url{http://arxiv.org/abs/1606.08415}
\BIBentrySTDinterwordspacing

\bibitem{agarap2019deeplearningusingrectified}
\BIBentryALTinterwordspacing
A.~F. Agarap, ``Deep learning using rectified linear units (relu),'' 2019. [Online]. Available: \url{https://arxiv.org/abs/1803.08375}
\BIBentrySTDinterwordspacing

\bibitem{ba2016layernormalization}
\BIBentryALTinterwordspacing
J.~L. Ba, J.~R. Kiros, and G.~E. Hinton, ``Layer normalization,'' 2016. [Online]. Available: \url{https://arxiv.org/abs/1607.06450}
\BIBentrySTDinterwordspacing

\bibitem{park2019specaugment}
D.~S. Park, W.~Chan, Y.~Zhang, C.-C. Chiu, B.~Zoph, E.~D. Cubuk, and Q.~V. Le, ``Specaugment: A simple data augmentation method for automatic speech recognition,'' in \emph{Proc. Interspeech}, 2019.

\bibitem{wu2021u2unifiedtwopassbidirectional}
\BIBentryALTinterwordspacing
D.~Wu, B.~Zhang, C.~Yang, Z.~Peng, W.~Xia, X.~Chen, and X.~Lei, ``U2++: Unified two-pass bidirectional end-to-end model for speech recognition,'' 2021. [Online]. Available: \url{https://arxiv.org/abs/2106.05642}
\BIBentrySTDinterwordspacing

\bibitem{vonneumann2023meeteval}
T.~von Neumann, C.~Boeddeker, M.~Delcroix, and R.~Haeb‑Umbach, ``{MeetEval}: A toolkit for computation of word error rates for meeting transcription systems,'' in \emph{Proceedings of the 7th International Workshop on Speech Processing in Everyday Environments (CHiME 2023)}, 2023, pp. 27--32.

\bibitem{meta_llama-3.2-1b-instruct}
{Meta}, ``Llama-3.2-1b-instruct,'' \url{https://huggingface.co/meta-llama/Llama-3.2-1B-Instruct}, 2024.

\bibitem{wang2023campp}
H.~Wang, S.~Zheng, Y.~Chen, L.~Cheng, and Q.~Chen, ``Cam++: A fast and efficient network for speaker verification using context-aware masking,'' in \emph{Proc. Interspeech 2023}, 2023.

\bibitem{phi4multimodal}
{Microsoft} and H.~F. community, ``Phi-4-multimodal-instruct: Lightweight open multimodal foundation model with text, image, and audio inputs,'' \url{https://huggingface.co/microsoft/Phi-4-multimodal-instruct}, 2025, 5.6B parameters.

\bibitem{loshchilov2017decoupled}
I.~Loshchilov and F.~Hutter, ``Decoupled weight decay regularization,'' \emph{arXiv preprint arXiv:1711.05101}, 2017.

\bibitem{kim2025phi4mm_ko_stt}
\BIBentryALTinterwordspacing
D.~Kim, ``{Phi‑4‑multimodal‑finetune‑KO‑speech}: A korean speech‑to‑text fine‑tuned model based on phi‑4 multimodal,'' Hugging Face model, May 2025, available at “daekeun‑ml/Phi‑4‑multimodal‑finetune‑ko‑speech” on Hugging Face. [Online]. Available: \url{https://huggingface.co/daekeun-ml/Phi-4-multimodal-finetune-ko-speech}
\BIBentrySTDinterwordspacing

\end{thebibliography}


\end{document}